\pdfoutput=1

\documentclass[11pt]{article}

\usepackage{acl}

\usepackage{times}
\usepackage{latexsym}
\usepackage[T1]{fontenc}

\usepackage[utf8]{inputenc}

\usepackage{microtype}

\usepackage{inconsolata}
\usepackage{amsmath}
\usepackage{txfonts}
\usepackage{boldline}
\usepackage{array}
\usepackage{graphicx}
\usepackage{tabularx}
\usepackage{arydshln}
\usepackage{multirow}
\usepackage{booktabs}
\newcolumntype{L}[1]{>{\arraybackslash}m{#1}}
\usepackage[export]{adjustbox}
\usepackage[most]{tcolorbox}
\usepackage{wasysym}
\usepackage{colortbl}
\usepackage[table]{xcolor}
\usepackage{enumitem}
\usepackage{subcaption}

\title{Finding Answers in Thought Matters:\\Revisiting Evaluation on Large Language Models with Reasoning}

\author{Hwiyeol Jo$^{1}$, Joosung Lee$^{1}$, Jaehong Lee$^{1}$, Sang-Woo Lee$^{2}$, Joonsuk Park$^{3}$$^\dagger$, Kang Min Yoo$^{1}$\thanks{\ \ Now at Amazon} \thanks{\ \ Co-corresponding author} \\ 
  $^1$NAVER Cloud, $^2$Neurofusion, $^3$University of Richmond \\
  \texttt{hwiyeolj@gmail.com,\{rung.joo,jaehong.l,kangmin.yoo\}@navercorp.com} \\
  \texttt{sam@neurofusion.ai,park@joonsuk.org}
}

\begin{document}
\maketitle
\begin{abstract}
    Evaluating generative models, such as large language models (LLMs), commonly involves question-answering tasks where the final answer is selected based on probability of answer choices. On the other hand, for models requiring reasoning, the method of answer extraction plays a critical role. Our research reveals that the performance of reasoning models and their final answer distributions are highly sensitive to the answer extraction algorithm employed. In order to mitigate this, we propose a basic framework: \textsc{Answer Regeneration}. The method uses an additional model inference, providing the prior input and output prefaced by the prompt "Answer:". The final answer is then selected or extracted from the regenerated output. We show that this extraction-rule-agnostic approach exhibits improved performance and enhanced robustness. Furthermore, we have applied this framework to general math problems and open-ended question answering tasks. Our analysis and this framework could offer a more reliable results for model evaluation.

\end{abstract}

\section{Introduction}\label{sec:intro}
    
    The conventional approach for generating answers from large language models (LLMs) involves selecting the answer choice with the highest probability when conditioned on the input prompt and each choice following a specific prefix, such as "Answer:" (\citet{hendryckstest2021,liang2023holisticevaluationlanguagemodels,2023opencompass,lighteval}; \textit{inter alia}). For tasks without answer choices, prior work has relied on rule-based extraction (e.g., searching for "Answer: X" or "answer is X"), model judges for semantic similarity, or human evaluation (\citet{kamalloo2023evaluating,wei2024measuring,chandak2025answer,chen2025xverify}; \textit{inter alia}).
    However, reasoning-powered LLMs
    need to output their reasoning process (Chain-of-Thought (CoT))~\citep{wei2022chain} to leverage their full potential. This detailed, linguistically diverse output complicates traditional evaluation. Specifically, it prevents the use of methods based on the probability of specific answer choices and limits the applicability of most LLM-as-a-judge~\citep{zheng2023judging} evaluations. This shift introduces a new, critical challenge: \textit{how to reliably find the answer from the detailed output that includes all the reasoning steps matters}.

    \begin{figure}[t]
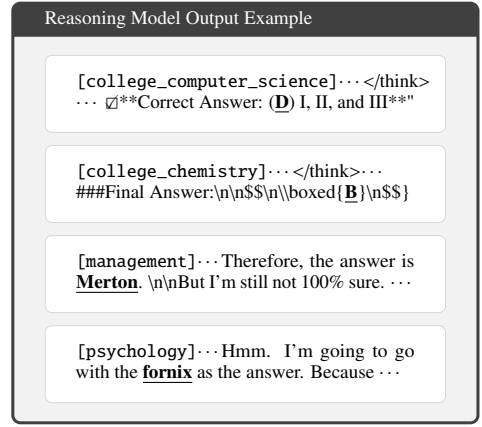
\centering
    \scalebox{.8}{
    \begin{tcolorbox}[colback=gray!10, colframe=black!75, title=\small Reasoning Model Output Example]\small
    \begin{tcbraster}[raster equal height, raster valign=top, raster columns=1, raster column skip=1mm, raster rows=2, colback=gray!10,colframe=black!75]
    \begin{tcolorbox}[boxrule=0.5pt,colback=white!50,colframe=gray!30]
        {\small\tt [college\_computer\_science]}{\small $\cdots$\textless /think\textgreater \\ $\cdots$ \CheckedBox **Correct Answer: (\underline{\bf D}) I, II, and III**"}
    \end{tcolorbox}
    \begin{tcolorbox}[boxrule=0.5pt,colback=white!50,colframe=gray!30]
        {\small\tt [college\_chemistry]}$\cdots$\textless /think\textgreater$\cdots$\\\#\#\#Final Answer:\textbackslash n\textbackslash n\$\$\textbackslash n\textbackslash \textbackslash boxed\{\underline{{\bf B}}\}\textbackslash n\$\$\}
    \end{tcolorbox}
    \begin{tcolorbox}[boxrule=0.5pt,colback=white!50,colframe=gray!30]
        {\small\tt [management]}$\cdots$Therefore, the answer is \underline{\bf Merton}. \textbackslash n\textbackslash nBut I'm still not 100\% sure. $\cdots$
    \end{tcolorbox}
    \begin{tcolorbox}[boxrule=0.5pt,colback=white!50,colframe=gray!30]
        {\small\tt [psychology]}{\small $\cdots$Hmm. I'm going to go with the \underline{\bf fornix} as the answer. Because $\cdots$}
    \end{tcolorbox}
    \end{tcbraster}
    \end{tcolorbox}
    }
    \caption{Examples illustrating the difficulties in extracting final answers from reasoning models' outputs. Although the benchmark is designed with multiple-choice questions, models frequently generate answers in a free-text format, which complicates automated evaluation.}\label{fig:intro_example}
    \vspace{-3mm}
    \end{figure}

    However, the rule-based approach suffers from a fundamental flaw: heuristic rules cannot account for all possible answer formats.
    Figure~\ref{fig:intro_example} illustrates examples
    from multiple-choice question answering benchmark MMLU~\cite{hendryckstest2021}. A single model can use different formats in its responses, sometimes boxing the answer in brackets (i.e., \textbackslash boxed\{\}) or answering the option text in various formats (e.g., "Merton", "fornix") instead of the option label (e.g., "(D)"). Furthermore, the formats can vary significantly between different models and even across different types of benchmarks, such as multiple-choice, math, and open-ended questions. This means that optimal extraction rules need to be created and tuned for every individual model and benchmark (e.g., rules for options, numbers, or word(s)), which makes the process difficult and even affects the reproducibility of model results.

    In this paper, we first empirically demonstrate the impact of answer extraction rules on reasoning-powered model (Section~\ref{sec:study1}). We then introduce \textsc{Answer Regeneration}, a simple, generation-based framework designed to alleviate the dependency on specific answer extraction rules (Section~\ref{sec:study2}). Instead of relying on complex extraction rules, our method utilizes an additional inference step to prompt the model to regenerate its final answer. It allows us to use probability-based answering for choices or extract the answer from a simplified output.
    
    Our experiments
    reveal that model performances are highly sensitive to the extraction rules employed. Depending on the rules, distinct answers---no answers at all in some cases---may be extracted from the same LLM response. On the other hand, \textsc{Answer Regeneration} consistently outperforms the handcrafted rule-based extractions, improving both in benchmark score and human evaluation results. Our method also achieves intuitive model rankings, where larger models are shown to outperform smaller ones. We demonstrate that \textsc{Answer Regeneration} significantly reduces the dependency on specific answer extraction rules, thereby improving robustness and reproducibility of model evaluations. Furthermore, we apply our framework to diverse tasks, including complex multiple-choice question answering, short-answer math problems, and open-ended question answering. In all cases, our generation-based method proves to be a plausible and effective approach for the fair evaluation of reasoning models.

    Our contributions in this work are as follows:

    \begin{itemize}[noitemsep,topsep=1pt]
        \item We empirically investigate the sensitivity of reasoning-powered LLMs to rule-based answering, revealing a strong dependency on the choice of answer extraction algorithm.
        \item We propose the generation-based framework \textsc{Answer Regeneration}. It achieves (1) superior performance compared with handcrafted rules, (2) intuitive model rankings, and (3) significantly enhanced robustness against answer inconsistency and incomplete outputs.
        \item We demonstrate the generalizability and effectiveness of our framework across diverse tasks, confirming its plausibility for more robust and fair model evaluations.
    \end{itemize}

\section{Related Work}

    A growing body of work shows that LLM performance can vary drastically with small changes in prompt format, even when the underlying semantics are equivalent~\citep{sclar2024quantifyinglanguagemodelssensitivity,he2024doespromptformattingimpact,alzahrani2024benchmarkstargetsrevealingsensitivity}. Consequently, \citet{polo2024efficientmultipromptevaluationllms,mizrahi-etal-2024-state} proposed the methods to mitigate the effect of prompt variations. While the previous research focused on \textit{input-level} prompt variations and their impact on model evaluation, we focus on \textit{output-level} final answer variations from reasoning LLMs, which are caused by the selection of answer extraction algorithms.

    Therefore, it is noteworthy to find out how recent LLM evaluations handle outputs from reasoning models. A number of open evaluation frameworks typically support (1) probability–based answering for multiple-choice tasks or (2) simple heuristic post-processing for free-form generations, involving only de-capitalization or blank-space normalization.
    Details on the implementations of {\bf MMLU Hendrycks}~\citep{hendryckstest2021}, {\bf HELM}~\citep{liang2023holisticevaluationlanguagemodels}, {\bf OpenCompass}~\citep{2023opencompass}, and {\bf lighteval}~\citep{lighteval} can be found in the Appendix~\ref{appendix:related_work}.
        
    {\bf lm-evaluation-harness}~\citep{biderman2024lessons} has become the de facto community standard for reproducible LLM evaluation. Generative tasks use string-match with optional regular expressions or rule-based normalizers. While recent templates support CoT prompting, the final answer is still recovered via simple patterns (e.g., "Answer: X"), or a last-capital-letter heuristic. As we will demonstrate, such extraction rules can swing scores and even reorder model rankings.

    To the best of our knowledge, this is the first study to highlight the importance of answer extraction methods, especially for reasoning LLMs.
    Based on our findings, we introduce a lightweight method to reduce the reliance on the fragile extraction rules and provides a more faithful evaluation of reasoning models' abilities.
    
    \begin{figure*}[t]\centering
        \vspace{-5mm}
        \includegraphics
        [width=0.84\textwidth,trim={1.2cm 0 0 0},clip]
        {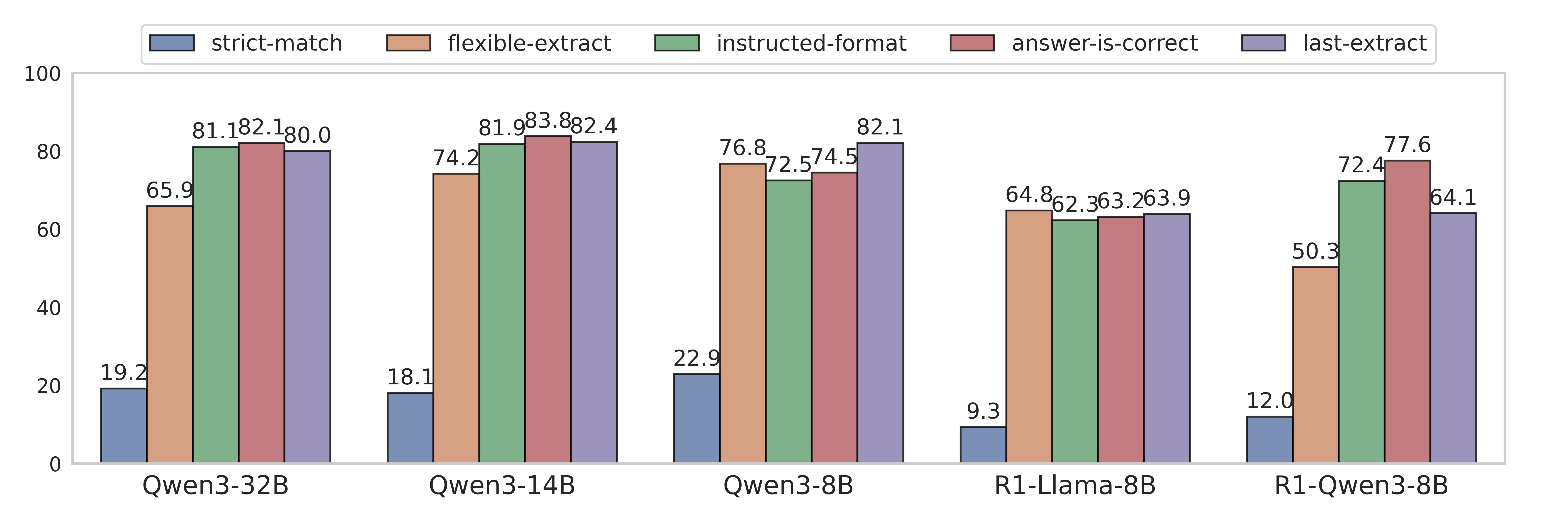}
        \vspace{-4mm}
        \caption{Model performance in accuracy evaluated using various answer extraction algorithm. Responses are considered incorrect if the extraction process fails to find an answer.}\label{fig:main_analysis_all}
        \vspace{-5mm}
    \end{figure*}
    
\section{Experiment Setup}

    The experiments are designed to highlight current problems associated with finding answers in reasoning models' output (Study 1 in Section~\ref{sec:study1}) and then assess the validity of introduced method \textsc{Answer Regeneration} (Study 2 in Section~\ref{sec:study2}).
    
    We utilize {\bf lm-evaluation-harness} toolkit for its simplicity in customizing the post-processing rules. {\bf MMLU}~\citep{hendryckstest2021} benchmark is primarily used, given its widely adoption for evaluating LLMs' knowledge\footnote{We select the original MMLU to better analyze how models handle ambiguous questions, rather than the cleaned MMLU-Redux~\citep{gema2025we}.}. The multiple-choice format of MMLU serves as a foundational task that simplifies the answer extraction process for our initial analysis. We then extend our evaluation to more complex tasks {\bf MMLU-Pro}~\citep{wang2024mmlu}, the mathematical reasoning benchmark {\bf GSM8K}~\citep{cobbe2021training} and the open-ended question answering {\bf TriviaQA}~\citep{joshi2017triviaqa} in Section~\ref{sec:further_study}.
    
    We evaluate several open-source reasoning models: Qwen3 families--{\bf Qwen3-32B}, {\bf Qwen3-14B}, {\bf Qwen3-8B}~\citep{yang2025qwen3}, along with {\bf Deepseek-R1-Distill-Llama-8B} (referred to as {\bf R1-Llama-8B}), and {\bf DeepSeek-R1-0528-Qwen3-8B} (referred to as {\bf R1-Qwen3-8B})~\citep{deepseekai2025deepseekr1incentivizingreasoningcapability}. 
    For hyperparameter settings, we adhere to recommended best practices for each model, setting temperature to 0.6, top-p value to 0.95, and top-k value to 20. Prompt templates are sourced from lm-evaluation-harness, using thinking templates. To reduce computational costs, the maximum token generation length is limited to 4,096.

    \begin{figure*}[t]\centering
    \scalebox{.75}{
    \begin{tcolorbox}[colback=gray!10, colframe=black!75, title=\small Answer Inconsistency Example]\small
        The following are multiple choice questions (with answers) about high school physics.
        
        Q: The plates of a capacitor are charged to a potential difference of 5 V. If the capacitance is 2 mF, what is the charge on the positive plate?
        
        (A) 0.005 C (B) 0.01 C (C) 0.02 C (D) 0.5 C
        
        $<$think$>$

        Okay, let's see. The question is about finding the charge on the positive plate of a capacitor. $\cdots$ the charges on the two plates are equal in magnitude but opposite in sign. So the positive plate has +Q and the negative has -Q. But the question is just asking for the magnitude, so the answer is still 0.01 C. Therefore, option \boxed{\text{{\tt answer-is-correct}: {\bf B}}} is correct.
        
        $<$/think$>$

        To determine the charge on the positive plate of a capacitor, $\cdots$ 
        Since the question asks for the **magnitude** of the charge on the **positive** plate, the correct answer is:
        \$\$
        \boxed{\text{{\tt instructed-format}: {\bf \textbackslash boxed\{0.01 \textbackslash , \textbackslash text\{C\}\}}}}
        \$\$
    
        ---

        \#\#\# Final Answer:

        **\boxed{\text{{\tt flexible-match}: {\bf (B)}}} 0.01 \boxed{\text{{\tt last-extract}: {\bf C}}}**
    \end{tcolorbox}
    }
    \caption{An example from Qwen3-32B output illustrating how the final answer can vary significantly depending on the extraction method used. The graphical boxes and bold text highlight the specific text extracted by each algorithm.}\label{fig:answer_inconsistency_example}
    \vspace{-3mm}
    \end{figure*}
    
\section{Study 1: Rule-based Answer Extraction}~\label{sec:study1}
    \vspace{-8mm}
    
    \subsection{Methods}
    We evaluate 5 reasoning models using 5 different answer extraction methods to investigate how performance changes with extraction algorithms: {\bf strict-match}, {\bf flexible-extract}, {\bf instructed-format}, {\bf answer-is-correct}, and {\bf last-extract}:

    {\bf strict-match} and {\bf flexible-extract} are adapted from lm-evaluation-harness. {\small\tt strict-match} extracts a precise string such as "answer is X" or "Answer: X" and {\small\tt flexible-extract}
    finds multiple-choice options like (A), (B), (C), or (D), located near the end of the text. This is a common and effective approach, as the final conclusion typically follows the reasoning. However, the original implementation has tendency to extract the last capital character from any text, which can lead to errors.

    {\bf instructed-format} requires modifying the input prompt to guide the model's output format. As recommended in Qwen3 technical report, we add a specific instruction to the prompt: "Please show your choice in the answer field with only the choice letter, e.g., "answer": "C"." Rules are then implemented to extract the answer from this specified format. While the method is strict (deviation is generally considered incorrect).
    
    Further heuristically optimized answer extraction methods are used: We build upon {\small\tt strict-match} by creating {\bf answer-is-correct}, which includes variations like "X is the answer" or "X could be the correct answer", addressing the limited scope. We also refine {\small\tt flexible-extract} by developing {\bf last-extract}, which specifically targets the last \textit{single} capital character that appears in the output. All the implementations accounts for minor variations like "**X**", "**Answer:** X" and are designed to select the last match to accommodate potential self-correction within the model's response.

    Our objective is not to declare any one extraction method superior. Instead, we aim to demonstrate the range of performance that can be achieved using widely-used, well-optimized, heuristic extraction algorithms on a given task. Consequently, this research shows that benchmark performance is not solely dependent on the reasoning model's ability but is significantly influenced by the chosen answer extraction rules. The exact regular expressions used are described in the Appendix~\ref{appendix:exact_regular_expression}. 
    
    \begin{figure*}[t]\centering
        \includegraphics
        [scale=0.37,trim={0.5cm 0 0 0},clip]
        {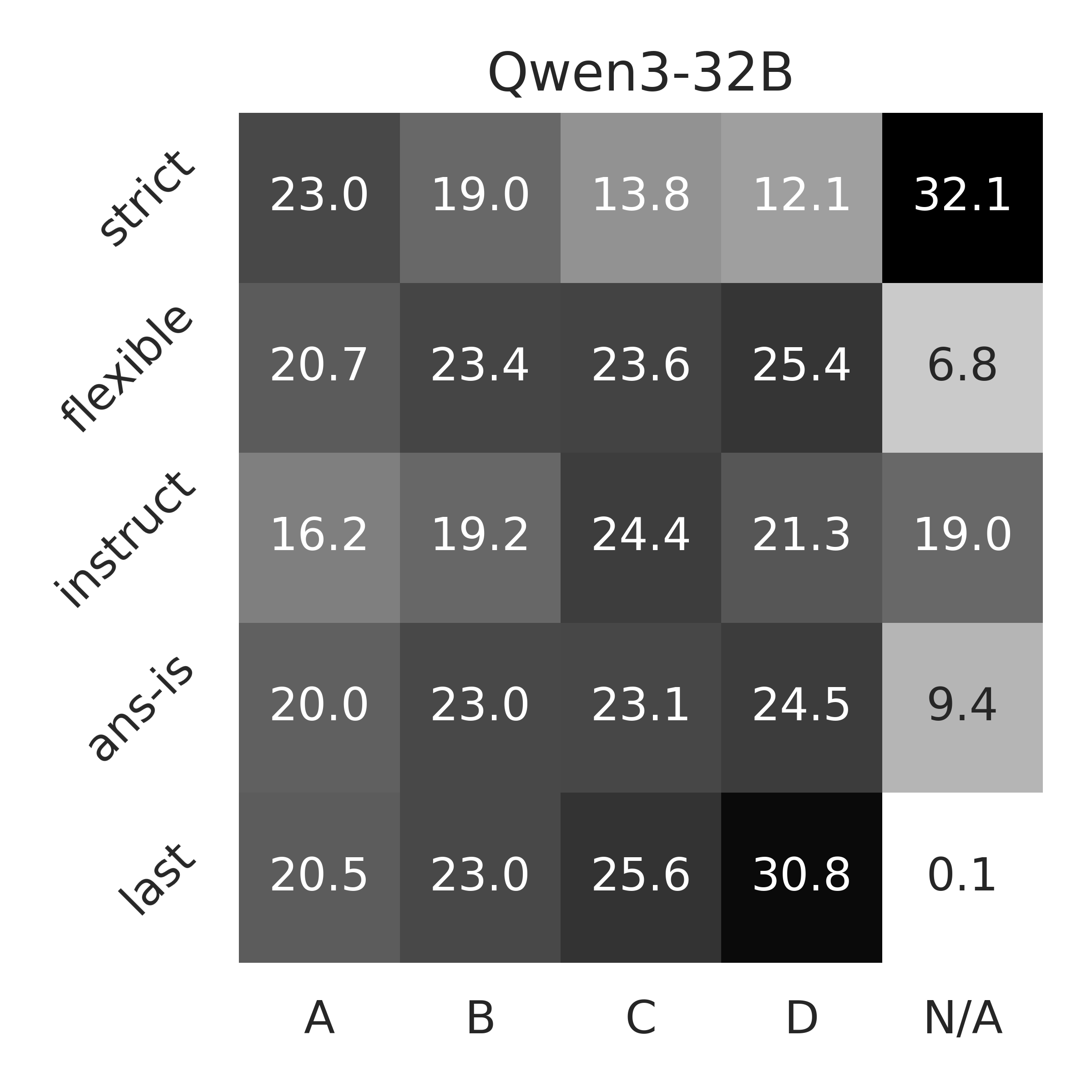}
        \includegraphics
        [scale=0.37,trim={2cm 0 0 0},clip]
        {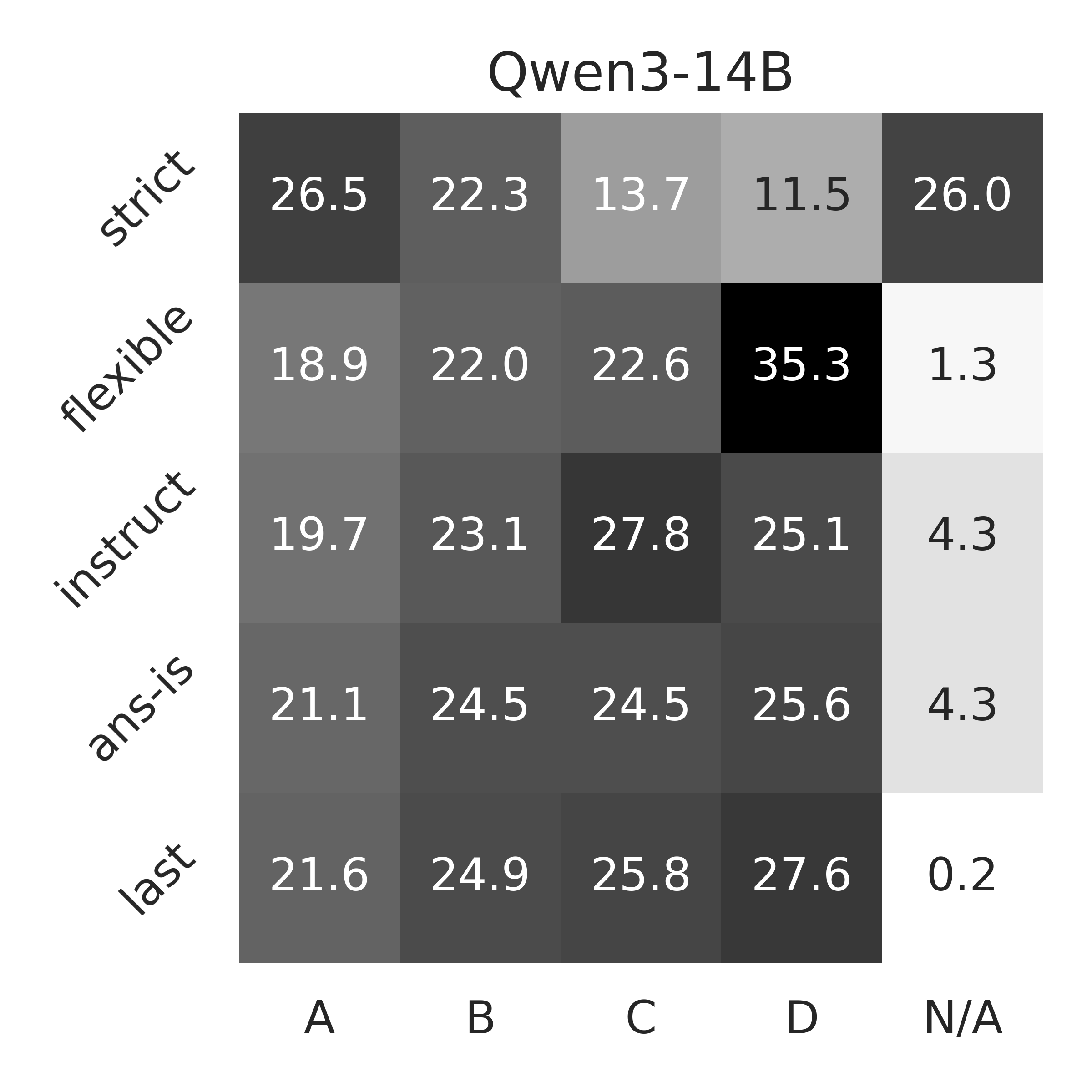}
        \includegraphics
        [scale=0.37,trim={2cm 0 0 0},clip]
        {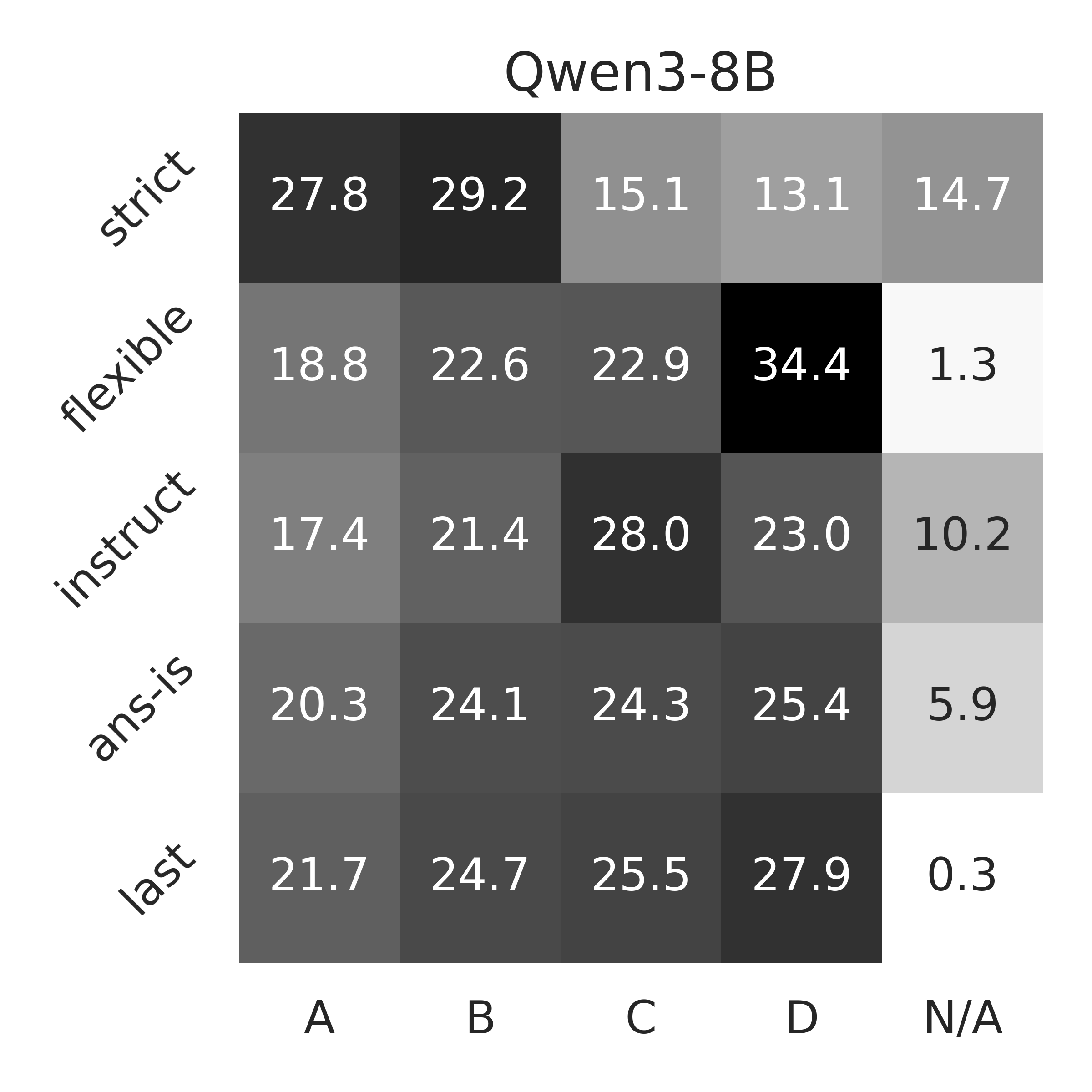}
        \includegraphics
        [scale=0.37,trim={2cm 0 0 0},clip]
        {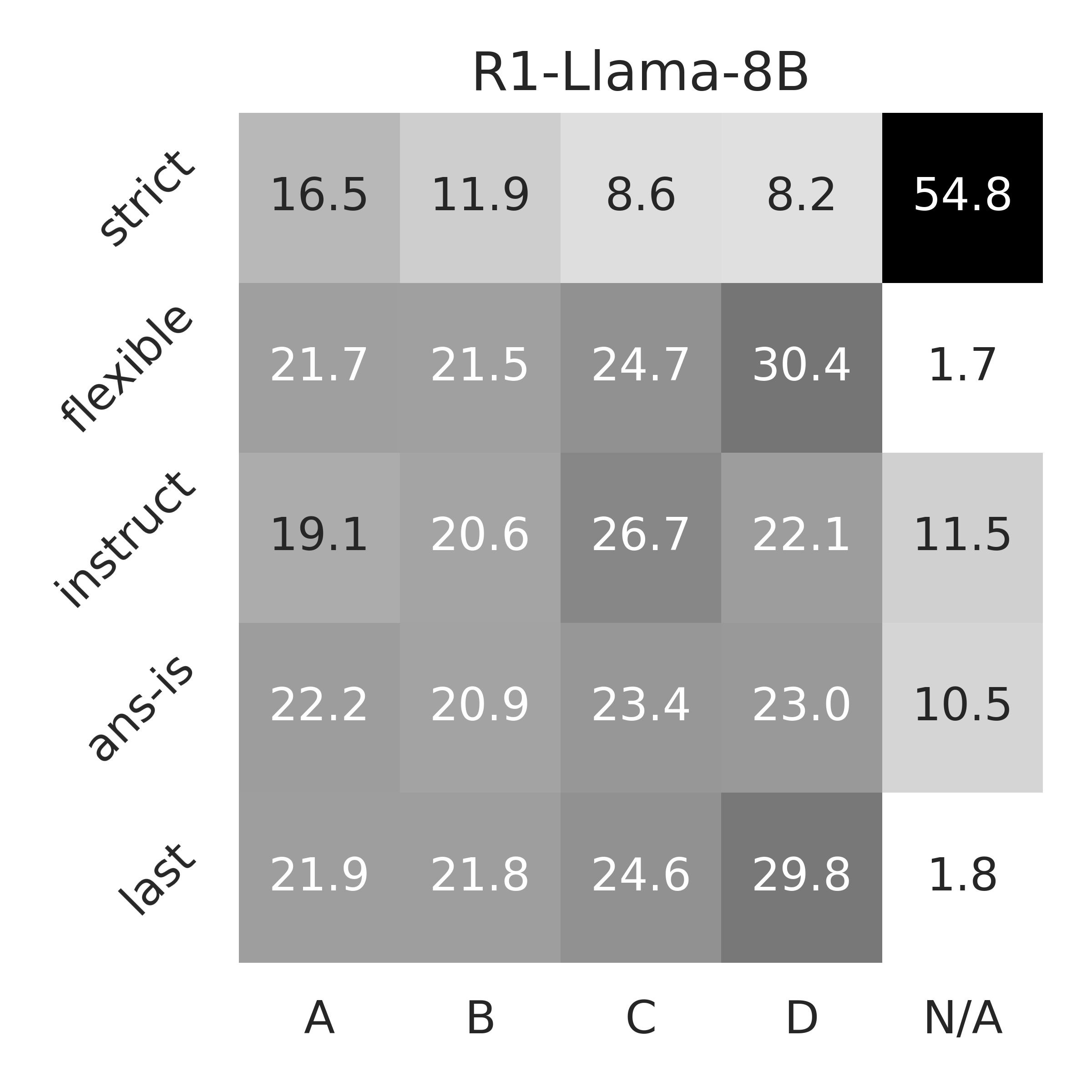}
        \includegraphics
        [scale=0.37,trim={2cm 0 0.5cm 0},clip]
        {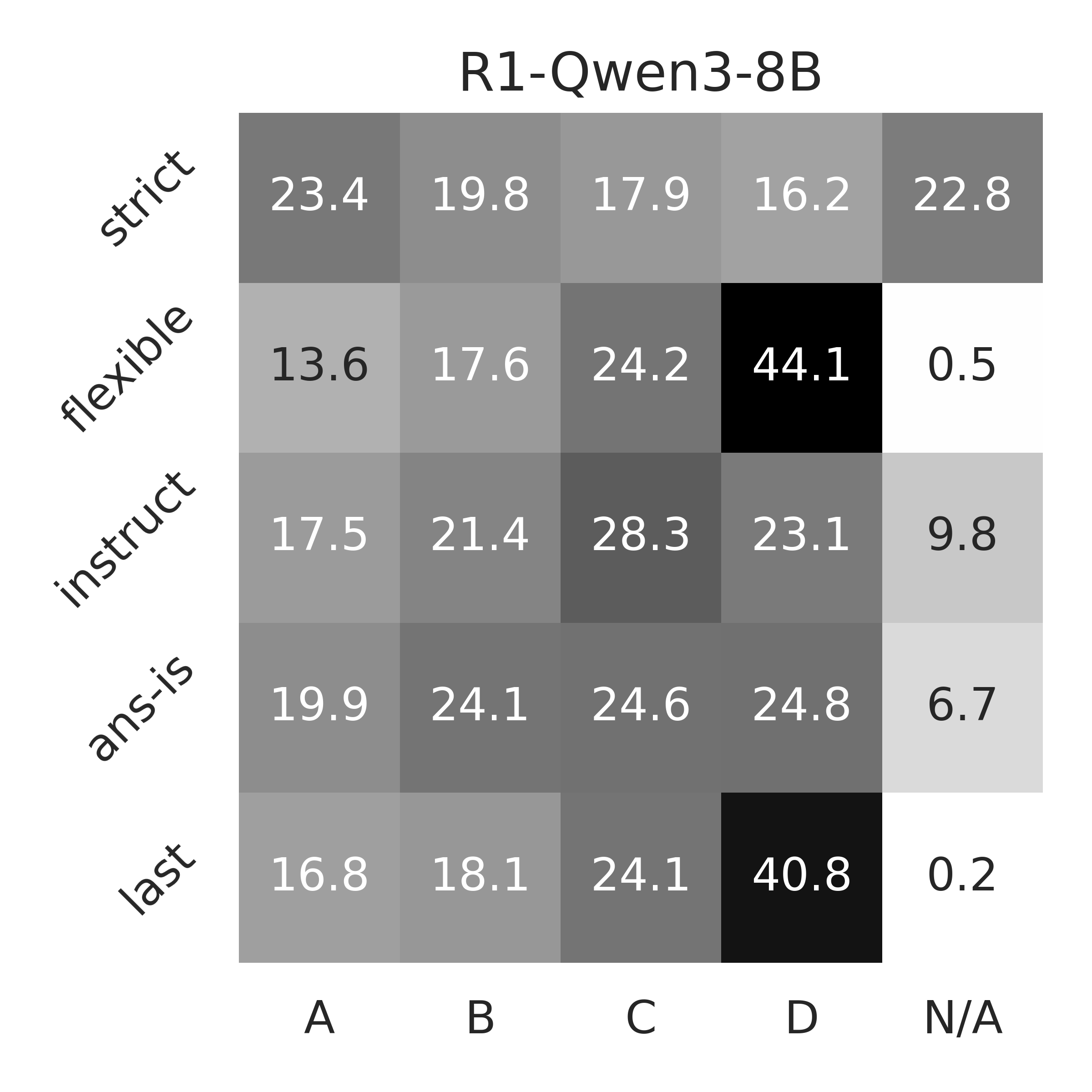}
        \vspace{-3mm}
        \caption{Distribution of extracted final answers across different extraction algorithms. The y-axis represents the answer extraction method, and the x-axis shows the extracted final answer, with "N/A" denoting cases where no answer could be extracted.}\label{fig:ans_dist}
        \vspace{-5mm}
    \end{figure*}

\subsection{Result}
\subsubsection{Model Performance}

    Figure~\ref{fig:main_analysis_all} illustrates how different answer extraction methods affect the performance of models. We evaluate performance using 3 types of rules: implemented ({\small\tt strict-match}, {\small\tt flexible-extract}), recommended ({\small\tt instructed-format}), and heuristically optimized ({\small\tt answer-is-correct}, {\small\tt last-extract}). If an extraction rule fails to find an answer, the response is considered incorrect. The results reveal that model performance fluctuates significantly depending on the extraction method used.

    With {\small\tt strict-match}, the rankings of model performances are Qwen3-8B, Qwen3-32B, Qwen3-14B, R1-Qwen3-8B, and R1-Llama-8B in order. The more optimized {\small\tt answer-is-correct}, derived from {\small\tt strict-match}, significantly improves the performance of all models. This shifts the ranking to Qwen3-14B, Qwen3-32B, Qwen3-8B, R1-Llama-8B, and R1-Qwen3-8B. A similar sensitivity is observed with the other methods. Using {\small\tt flexible-extract}, the top models are Qwen-8B, Qwen3-14B, Qwen-32B, R1-Llama-8B, and R1-Qwen3-8B. With {\small\tt last-extract}, Qwen3-14B performs the best, and R1-Qwen3-8B outperforms R1-Llama-8B compared with {\small\tt flexible-extract}. Interestingly, despite following the recommended best practices for multiple-choice question answering with {\small\tt instructed-format}, the performance of Qwen3 family models are not impressive compared to other extraction methods. This method proves to be particularly ineffective for R1-Llama-8B model.

    These findings challenge the common assumption that larger models outperform smaller ones within the same family. Our analysis indicates that the benchmark performance scores of reasoning models are highly dependent on the answer extraction method used. This suggests that the discrepancies between publicly reported and reproduced performance scores may be due to differences not only in prompt inputs, but also in the specific answer extraction methods, which are not fully disclosed.

    \begin{figure*}[t]\centering
    \includegraphics
        [width=0.8\linewidth]
        {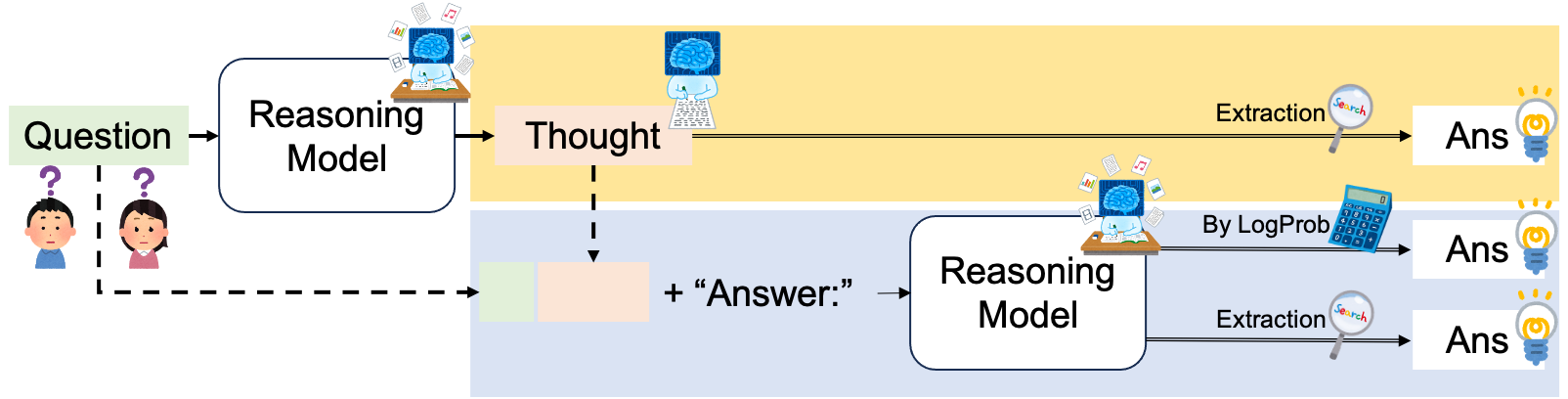}
    \caption{The proposed \textsc{Answer Regeneration} framework for finding answers in model output. The yellow box indicates the conventional method of direct extraction, while the blue box indicates the proposed framework.}\label{fig:proposed_result}
    \vspace{-5mm}
    \end{figure*}
    
\subsubsection{Answer Inconsistency}\label{sec:answer_inconsistency}
    Figure~\ref{fig:answer_inconsistency_example} provides a clear example of how different extraction methods handle the same model output, illustrating the problem of answer inconsistency. In this example, {\small\tt strict-match} fails. {\small\tt answer-is-correct} successfully locates an answer within the model's thought, between <think> and </think> tags. However, the model's explicitly formatted final answer, "the correct answer is: X", is not recognized as a valid because it contains unexpected patterns, including \$\$, \textbackslash boxed\{\}, and \textbackslash text\{\}. Besides, the output is the option text rather than the required option label. {\small\tt instructed-format} could find an answer using \textbackslash boxed\{\} (despite the \textbackslash boxed\{\} format being recommended only for math problems), but the extracted answer is again the option text, not the label. Furthermore, the presence of the unexpected LaTeX command \textbackslash text\{\} could result in an incorrect evaluation during string-match comparison. Meanwhile, {\small\tt flexible-match} correctly identifies the final answer. Interestingly, the simple yet effective {\small\tt last-extract} extracts the unit of option text "C" as the final answer.

    Figure~\ref{fig:ans_dist} further illustrates this issue by showing how the distribution of extracted answers changes depending on the extraction method used.
    We observe that the distribution of extracted answers varies significantly. This highlights the crucial role of the extraction method in determining model's final performance, suggesting that the choice of method can introduce bias into the evaluation.

    \begin{table}[t]\centering\small
    \scalebox{.80}{
        \begin{tabular}{@{}
        l@{\hskip 0.18cm}
        p{1.4cm}@{\hskip 0.18cm}p{1.4cm}@{\hskip 0.18cm}p{1.4cm}@{\hskip 0.18cm}p{1.3cm}@{\hskip 0.18cm}p{1.3cm}}
        \toprule
        & {\tt Qwen3-32B} & {\tt Qwen3-14B} & {\tt Qwen3-8B} & {\tt R1-Llama} & {\tt R1-Qwen3} \\
        \midrule
        {\tt (\%)} & 2.8 & 2.9 & 6.2 & 6.7 & 6.8 \\
        \midrule
        {\tt best-extr} & {\tt ans-is} & {\tt ans-is} & {\tt last} & {\tt flexible} & {\tt ans-is}\\
        \midrule
        {\tt Correct} & 37.1 & 33.8 & 42.1 & 26.6 & 25.5 \\
        {\tt Incorrect} & 32.2 & 22.6 & 53.6 & 65.7 & 19.2 \\
        {\tt Invalid} & 30.7 & 43.6 & 4.4 & 7.8 & 55.3 \\
        \bottomrule
        \end{tabular}
    }
    \caption{The percentage of incomplete thinking and the corresponding accuracy of each reasoning model. (\texttt{\%}) refers to the portion of outputs where model's thinking process is not completed.
    }\label{tab:failure-case}
    \vspace{-7mm}
    \end{table}
    
\subsubsection{Answering for Incomplete Thinking}\label{sec:endless_thinking}

    Another challenge in extracting answers from reasoning models is the issue of incomplete reasoning (or thinking). Even when we set the maximum generation length to 4,096 tokens, we find that some model outputs lack the </think> token, indicating that the thinking process had not concluded. Table~\ref{tab:failure-case} reports the percentage of outputs in this category. Fortunately, this is a relatively small portion of the total outputs and is primarily caused by repetitions during the model's generation.

    We then select the best answer extraction method for each model and measure the correctness of the final answers derived from these incomplete outputs. Except for Qwen3-8B and R1-Llama-8B, which use extraction algorithms solely on capital letters, the results using {\small\tt answer-is-correct} show a high rate of invalid extraction. This implies that even well-optimized extraction method can be less robust toward incomplete thinking, particularly when the reasoning output does not contain definitive, explicitly formatted answering text.

    \begin{figure*}[t]\centering
        \vspace{-5mm}
        \includegraphics[scale=0.65, valign=c,trim={0.0cm 0.5cm 3.0cm 0.0cm},clip]{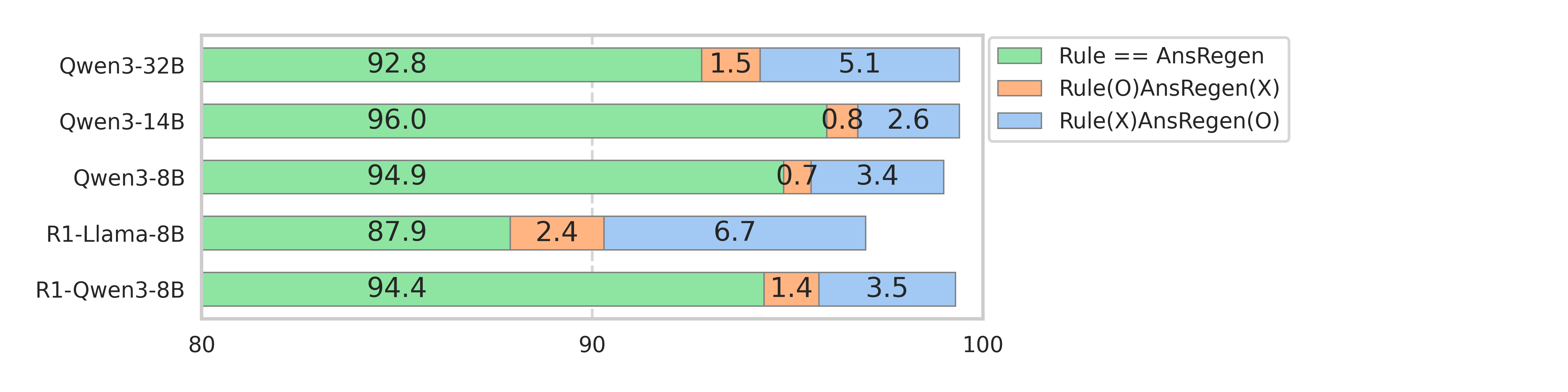}
        \hspace{-3mm}
        \includegraphics[scale=0.5, valign=c,trim={1.0cm 0.5cm 0.0cm 0.0cm},clip]{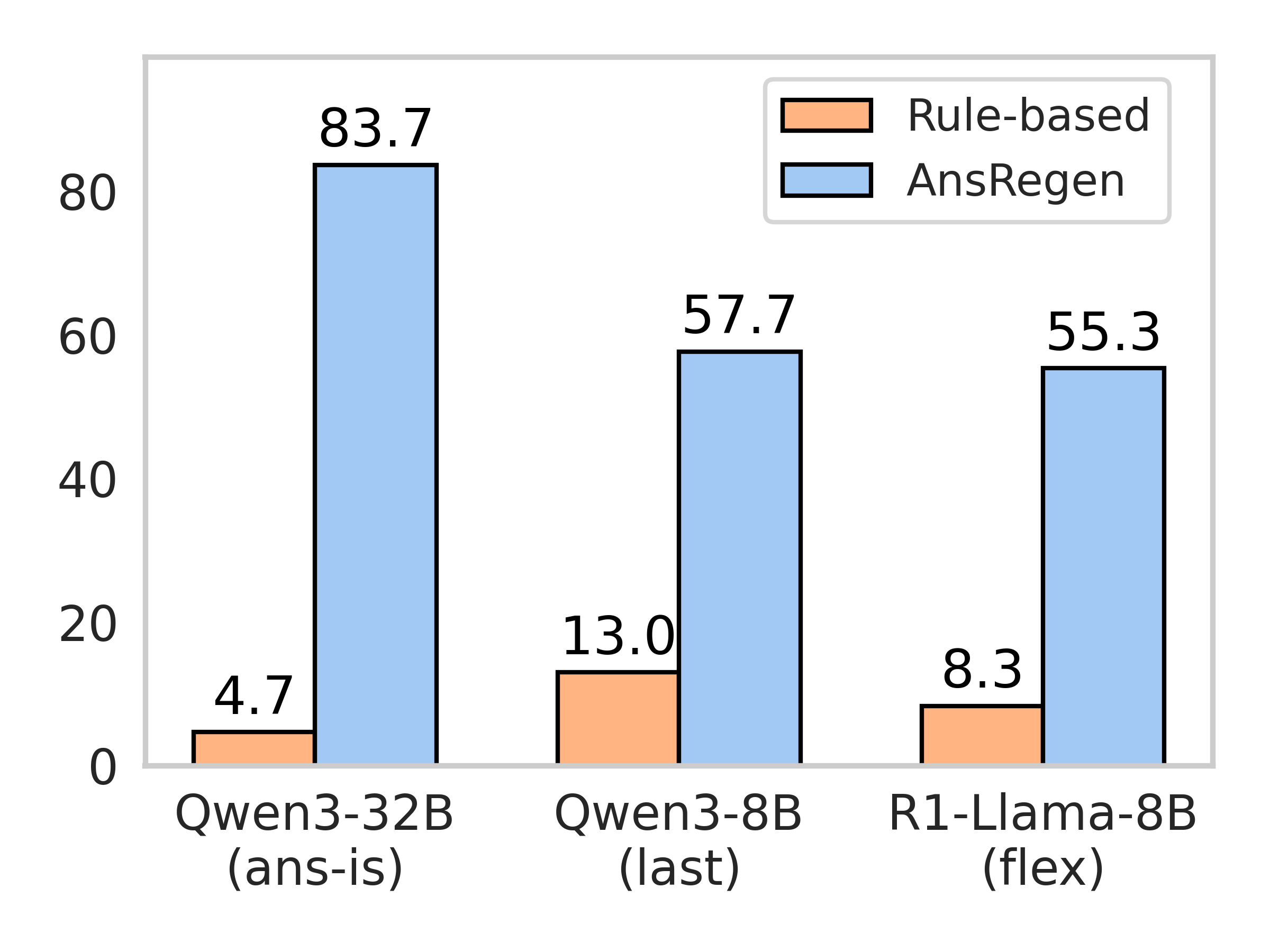}
        \vspace{-3mm}
        \caption{(left) A confusion matrix comparing the conventional answer extraction method ({\small\tt Rule}) and the proposed method ({\small\tt Regen}). (right) The accuracy of answers extracted from the model's thought, as determined by human evaluation. We sample 300 instances when the extraction and regeneration are disagreed. Results are not reported for cases where the model failed to provide a definitive answer or provided multiple option labels.
        }\label{fig:proposed_ablation}
        \vspace{-5mm}
    \end{figure*}
    
\section{Study 2: Answer Generation}~\label{sec:study2}
    \vspace{-5mm}
    
    Our analysis has shown that the final answer of reasoning models is highly sensitive to the chosen extraction method. Model performance fluctuates significantly based on how the answer is located and selected from the output. To address this and simplify the optimization of complex extraction algorithms, we propose a straightforward framework for reliably identifying the final answer.

\subsection{Method}\label{sec:method}

    Our proposed framework, illustrated in Figure~\ref{fig:proposed_result}, tackles the challenge by introducing \textsc{Answer Regeneration} step. Instead of attempting to parse a final answer from model's extensive thought, our method uses an additional inference call. Specifically, we provide the model (in its non-reasoning mode) with the original input prompt and its previous output (the reasoning process), and a new prefix "Answer:". This prompts the model to generate a concise, final answer based on its prior reasoning.

    This approach offers key benefits. For multiple-choice questions, it allows us to utilize probability-based answering, as non-reasoning models have been evaluated, leading to more robust predictions. When the answer choices are not available, such as open-ended question answering, it simplifies the model's output, making the final answer much easier to extract with straightforward algorithms.

    While effective, our framework has several acknowledged limitations. The primary issue is the computational cost of the additional inference step.
    Additionally, the method might not fully capture minor variations in answer formatting, e.g., the probability of "**A**". Finally, some regenerated results could be different from explicitly mentioned answer.
    Despite these weaknesses and the lack of technical novelty, we believe this framework's simplicity and the clarity constitute significant contribution. We will demonstrate its benefits using the same experimental setup as our previous analyses.

\subsection{Result}
    
    \begin{table}[t]\centering\small
    \scalebox{.82}{
        \begin{tabular}{@{}
        l@{\hskip 0.18cm}
        p{1.4cm}@{\hskip 0.18cm}p{1.4cm}@{\hskip 0.18cm}p{1.4cm}@{\hskip 0.18cm}p{1.2cm}@{\hskip 0.18cm}p{1.2cm}}
        \toprule
        & {\tt Qwen3-32B} & {\tt Qwen3-14B} & {\tt Qwen3-8B} & {\tt R1-Llama} & {\tt R1-Qwen3} \\
        \midrule
        {\tt Rule(Best)} & 82.1 & 83.8 & 82.1 & 64.8 & 77.6 \\
        {\tt AnsRegen} & \bf 87.1 & \bf 85.0 & \bf 83.3 & \bf 68.8 & \bf 80.7 \\
        \midrule
        {\tt Diff} & +5.0 & +1.2 & +1.2 & +4.0 & +3.1 \\
        \bottomrule
        \end{tabular}
    }
    \caption{Performance comparison between conventional answer extraction and \textsc{Answer Regeneration}. We report each model's performance using its best-performing extraction method.}\label{tab:proposed_performance}
    \vspace{-5mm}
    \end{table}
    
\subsubsection{Improved Performance}

    As presented in Table~\ref{tab:proposed_performance}, the proposed method consistently reports better scores than rule-based answering.
    Figure~\ref{fig:proposed_ablation} (left) provides a detailed look at the performance. While most of the final answers derived by both our method and the rule-based methods are the same, our framework achieves a much higher \textit{correction rate}. This demonstrates \textsc{Answer Regeneration} is successful at correcting incorrect answers extracted by rule-based approach.

    To compute the correction rate, we select 300 instances from the outputs of Qwen3-32B, Qwen3-8B, and R1-Llama-8B where the extraction and regeneration results disagreed. We then manually label the correct "gold" answers in terms of answer extraction from the thoughts. As shown in Figure~\ref{fig:proposed_ablation} (right), the agreement rate of \textsc{Answer Regeneration} with the human label is far superior to that of the conventional answer extraction methods.

    \subsubsection{Correlation with Model Size}
    
    An interesting effect of our framework is the change in the performance ranking of Qwen3 models. The previous ranking derived from rule-based answering, which was Qwen3-14B, Qwen3-32B, Qwen3-8B, shifted to Qwen3-32B, Qwen3-14B, Qwen3-8B under our framework. This new ranking aligns with conventional intuition and general knowledge that larger models typically outperform smaller ones within the same family. This suggests that the initial, counterintuitive ranking is likely an artifact of the answer extraction methods, not a true reflection of the models' underlying capabilities.

    \begin{figure}[t]\centering
    \vspace{-2mm}
        \includegraphics
        [width=0.75\linewidth,trim={1.0cm 0 0 0.0cm},clip]{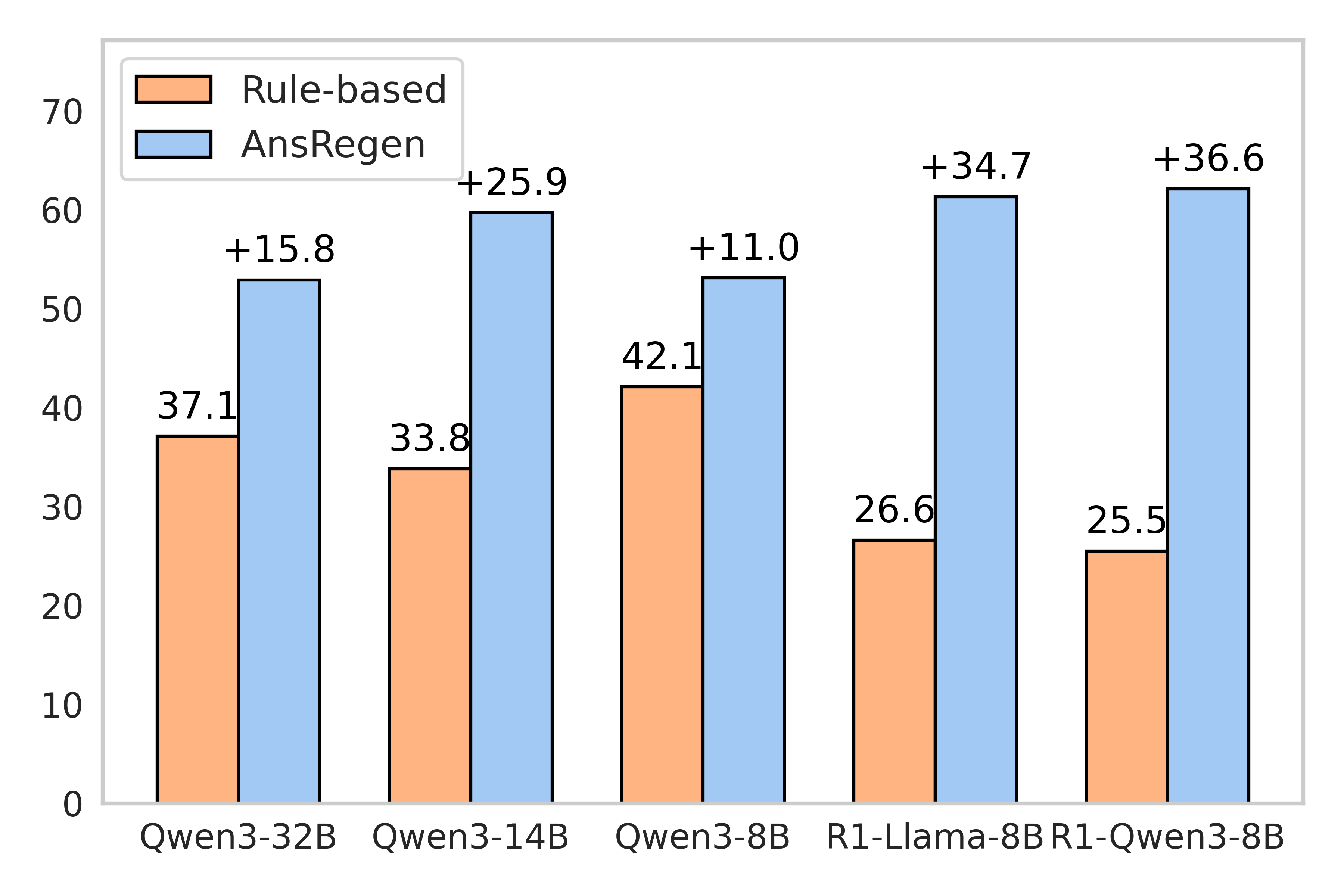}       
    \vspace{-3mm}
    \caption{Model performance evaluated on outputs where the reasoning process is incomplete, using the optimal answer extraction algorithm for each model.}\label{fig:incomplete_thinking_correction}
    \vspace{-5mm}
    \end{figure}
    
\subsubsection{Enhanced Robustness to Responses}

    The nature of our proposed \textsc{Answer Regeneration} framework inherently addresses the issue of answer inconsistency mentioned in Section~\ref{sec:answer_inconsistency}. Since it prompts the model to generate a final, definitive answer, it bypasses the unpredictable results associated with various rule-based extraction algorithms.

    Additionally, our method improves robustness by handling internal self-correction within model outputs. When facing ambiguous questions, a model may initially provide an answer and then continue its thinking process, generating alternative solutions or re-evaluating its answer. Rule-based answer extraction methods struggle to choose the final answer from this internal debate. In contrast, our framework considers the entire thinking process and forces the model to finalize its response, leading to a more reliable result.

    A further key advantage is its ability to handle "NOT correct" questions. Since many extraction algorithms are designed to find the "correct" answer from the reasoning text, they fail when the question requires identifying the incorrect choice. The algorithm may mistakenly extract a correct option discussed during the model's rumination.
    
    Finally, our method significantly improves performance in cases of incomplete thinking, as shown in Figure~\ref{fig:incomplete_thinking_correction}. Instead of relying on rules to parse an incomplete output, our framework can select the final answer even when the thought does not include an explicit final answer.

    \begin{table}[t]\centering\small
    \scalebox{.78}{
        \begin{tabular}{@{}
        l@{\hskip 0.18cm}
        p{1.4cm}@{\hskip 0.18cm}p{1.4cm}@{\hskip 0.18cm}p{1.4cm}@{\hskip 0.18cm}p{1.2cm}@{\hskip 0.18cm}p{1.2cm}}
        \toprule
        & {\tt Qwen3-32B} & {\tt Qwen3-14B} & {\tt Qwen3-8B} & {\tt R1-Llama} & {\tt R1-Qwen3} \\
        \midrule
        {\tt strict-match} & 15.3 & 13.0 & 15.7 & 6.8 & 10.9 \\
        {\tt flexible-ext} & 47.2 & 47.1 & 47.1 & 38.0 & 41.3 \\
        {\tt instructed} & 52.6 & 59.5 & 45.8 & 38.7 & 49.7 \\
        {\tt ans-is-corr} & 68.4 & 65.2 & 64.2 & 37.6 & 53.5 \\
        {\tt last-extract} & 66.8 & 63.4 & 62.0 & 42.2 & 45.3 \\
        {\tt implemented} & \bf 72.1 & \bf 69.4 & \bf 64.6 & \bf 43.3 & \bf 58.3 \\
        \midrule
        {\tt AnsRegen} & \bf 77.0 & \bf 72.6 & \bf 72.0 & \bf 43.6 & \bf 66.4 \\
        \midrule
        {\tt Reported}$^2$ & 79.8 & 77.4 & 74.3 & 54.3 & 73.9 \\
        {\tt $\triangleright$ Reproduced} & 63.0 & 59.2 & 57.3 & 42.3 & 40.7 \\
        \bottomrule
        \end{tabular}
    }
    \caption{Model performance on MMLU-Pro. The evaluation utilizes the same answer extraction algorithms used in our MMLU analysis, including the built-in algorithm from lm-evaluation-harness, referred to as \texttt{\small implemented}.}\label{tab:mmlu_pro_result}
    \vspace{-5mm}
    \end{table}
    
\subsubsection{Regenerator Independency}

    Our method, which uses an additional model inference for \textsc{Answer Regeneration}, raises a question about its dependency on the specific model used.

    Table~\ref{tab:llm-as-judge} in the Appendix shows that the performances achieved using small-sized regenerators are generally similar to the performance achieved when using the same model both for reasoning and \textsc{Answer Regeneration}. While this suggests a degree of independence, we still recommend using the same model for both tasks. The final answering step is also a crucial part of the overall model evaluation and should be performed by the model being assessed to ensure a consistent and fair comparison.
    
    Based on the results, \textsc{Answer Regeneration} framework shows a more effective and reliable method for evaluating reasoning models. Conventional extraction rules cannot account for all the variations in model outputs, and can thus introduce biases and inaccuracies. Our framework mitigates this problem, providing a more accurate and consistent measure of models' true performance.

\section{Studies on Additional Tasks}~\label{sec:further_study}
\vspace{-8mm}

\subsection{Complex Multiple-Choice Question Answering}

    As an extension of our previous findings, we investigate our framework on MMLU-Pro~\citep{wang2024mmlu}, a more complex benchmark with a dynamic number of answer options. The result, shown in Table~\ref{tab:mmlu_pro_result}, demonstrates that while the built-in extraction algorithm from lm-evaluation-harness performs better than algorithms optimized only for the original MMLU, \textsc{Answer Regeneration}—which is not specifically tuned for any benchmark—still achieves superior performance. Furthermore, the scores are also closer to the publicly reported performance\footnote{
    \url{https://artificialanalysis.ai/evaluations/mmlu-pro}.
    Qwen3 technical report does not contain zero-shot CoT results for MMLU-Pro; it only provides 5-shot results without reasoning, scoring 65.54 for 32B and 56.73 for 8B.}, despite the reported scores likely benefiting from more specific prompt engineering (e.g., detailed task descriptions for individual subtasks), as demonstrated in our reproduced score using their extraction rules. Therefore, we believe that evaluating models with our framework provides a fairer and more robust assessment of true capabilities, achieving competitive performance without the need for task-specific optimization.

    \begin{table}[t]\centering\small
    \scalebox{.78}{
        \begin{tabular}{@{}
        l@{\hskip 0.18cm}
        p{1.4cm}@{\hskip 0.18cm}p{1.4cm}@{\hskip 0.18cm}p{1.4cm}@{\hskip 0.18cm}p{1.2cm}@{\hskip 0.18cm}p{1.2cm}}
        \toprule
        ($\downarrow$) Extraction & {\tt Qwen3-32B} & {\tt Qwen3-14B} & {\tt Qwen3-8B} & {\tt R1-Llama} & {\tt R1-Qwen3} \\
        \midrule
        {\tt strict-match} & 3.3 & 2.7 & 1.7 & 0.0 & 0.1 \\
        {\tt flexible-ext} & 33.3 & 33.5 & 19.3 & 69.2 & 85.1 \\
        {\tt instructed} & \bf 93.5 & \bf 92.2 & 88.6 & 54.8 & \bf 85.8 \\ 
        {\tt ans-is-corr} & 89.6 & 87.6 & \bf 91.9 & \bf 63.1 & 83.4 \\
        \midrule
        {\tt AnsRegen} & \bf 95.0 & \bf 93.8 & 91.1 & \bf 76.0 & \bf 91.1 \\
        \bottomrule
        \end{tabular}
    }
    \caption{Model performance on GSM8K. Note that {\small\tt strict-match} and {\small\tt flexible-extract} are implemented in lm-evaluation-harness. {\small\tt last-extract} is not useful.}\label{tab:gsm8k_result}
    \vspace{-5mm}
    \end{table}

    \begin{table*}[t]\centering\small
    \begin{adjustbox}{scale=.75}
        \begin{tabular}{@{}
        lp{1.0cm}l@{\hskip 0.18cm}
        p{1.4cm}@{\hskip 0.18cm}p{1.4cm}@{\hskip 0.18cm}p{1.4cm}@{\hskip 0.18cm}p{1.4cm}@{\hskip 0.18cm}p{1.4cm}}
        \toprule
         & {\tt Method} & ($\downarrow$) {\tt Evaluator} & {\tt Qwen3-32B} & {\tt Qwen3-14B} & {\tt Qwen3-8B} & {\tt R1-Llama} & {\tt R1-Qwen3} \\
        \midrule
        \multirow{2}{*}{\shortstack{\tt String\\\tt Match}} & \tt ans-is-corr & \ \ \ \ \ - & 42.7 & 47.5 & 44.2 & 11.7 & 35.6 \\
        & \tt AnsRegen & \ \ \ \ \ - & 55.3$^*$ & 53.7 & 47.0 & 24.1 & 55.8 \\
        \midrule
        \multirow{7}{*}{\shortstack{\tt Model\\\tt-based}} & \multirow{5}{*}{\shortstack{\tt GPT\\\tt Grader}} &  {\tt Qwen3-32B} & 3.1$^*$ & 3.8 & 3.6 & 1.3 & 2.9 \\
        & & {\tt Qwen3-14B} & 49.5 & 56.4 & 49.2 & 19.4 & 41.1 \\
        & & {\tt Qwen3-8B} & 49.4 & 56.3 & 49.4 & 18.2 & 41.3 \\
        & & {\tt R1-Llama-8B} & 93.9 & 92.3 & 89.4 & 93.1$^*$ & 87.5 \\
        & & {\tt R1-Qwen3-8B} & 47.6 & 54.8 & 47.9 & 17.7 & 39.6 \\
        \cline{2-8}
        & {\tt xVerify} & {\tt xVerify-8B-I} & 0.0 & 0.0 & 0.0 & 47.7$^*$ & 0.0 \\
        \bottomrule
        \end{tabular}
    \end{adjustbox}
    
    \begin{subfigure}{0.35\linewidth}\centering
    \includegraphics[scale=0.42, valign=t,trim={0.5cm 1.0cm 0.25cm 0.25cm},clip]{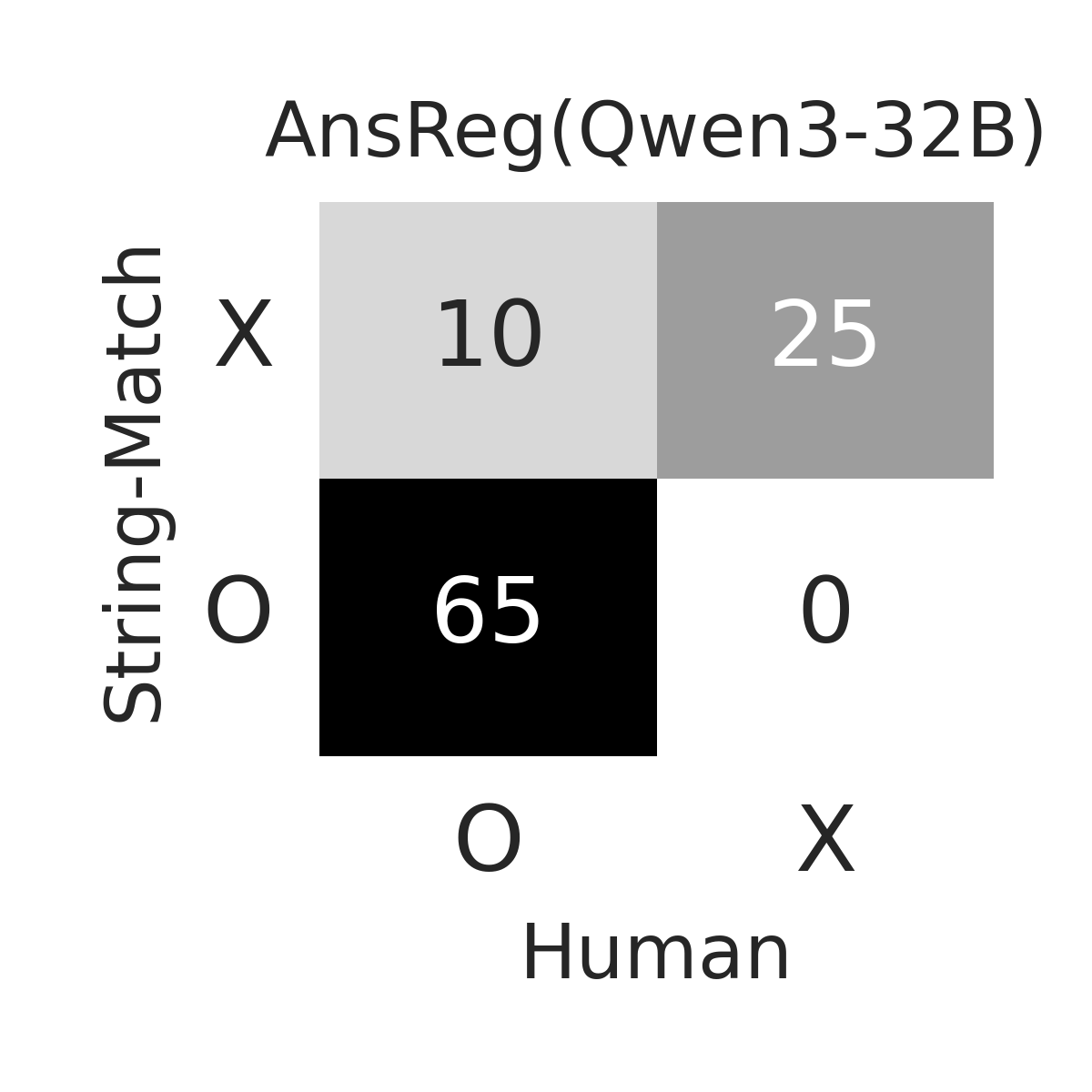}
    \includegraphics[scale=0.42, valign=t,trim={0.5cm 1.0cm 0.25cm 0.25cm},clip]{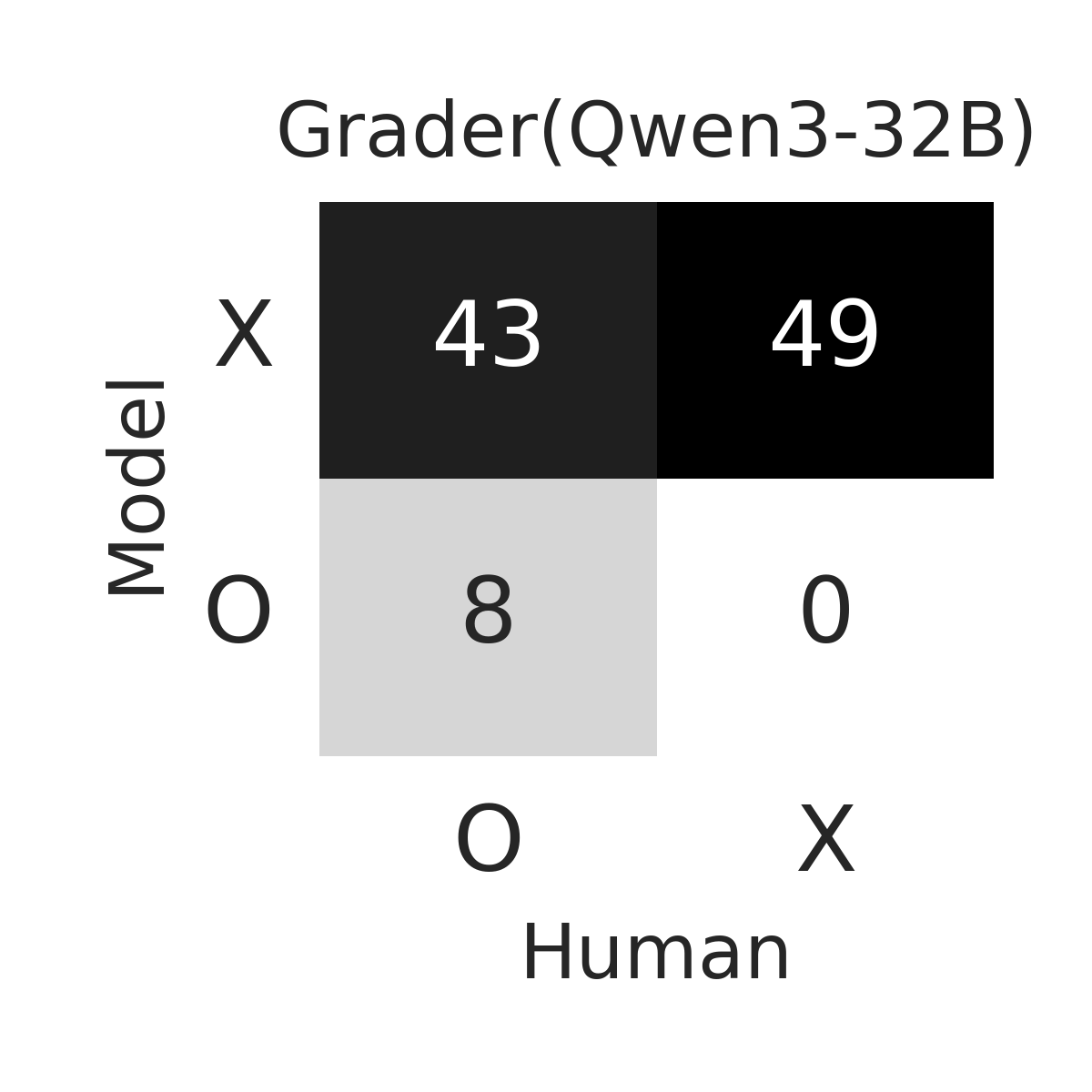}
    \\
    \includegraphics[scale=0.42, valign=t,trim={0.5cm 0.5cm 0.25cm 0.25cm},clip]{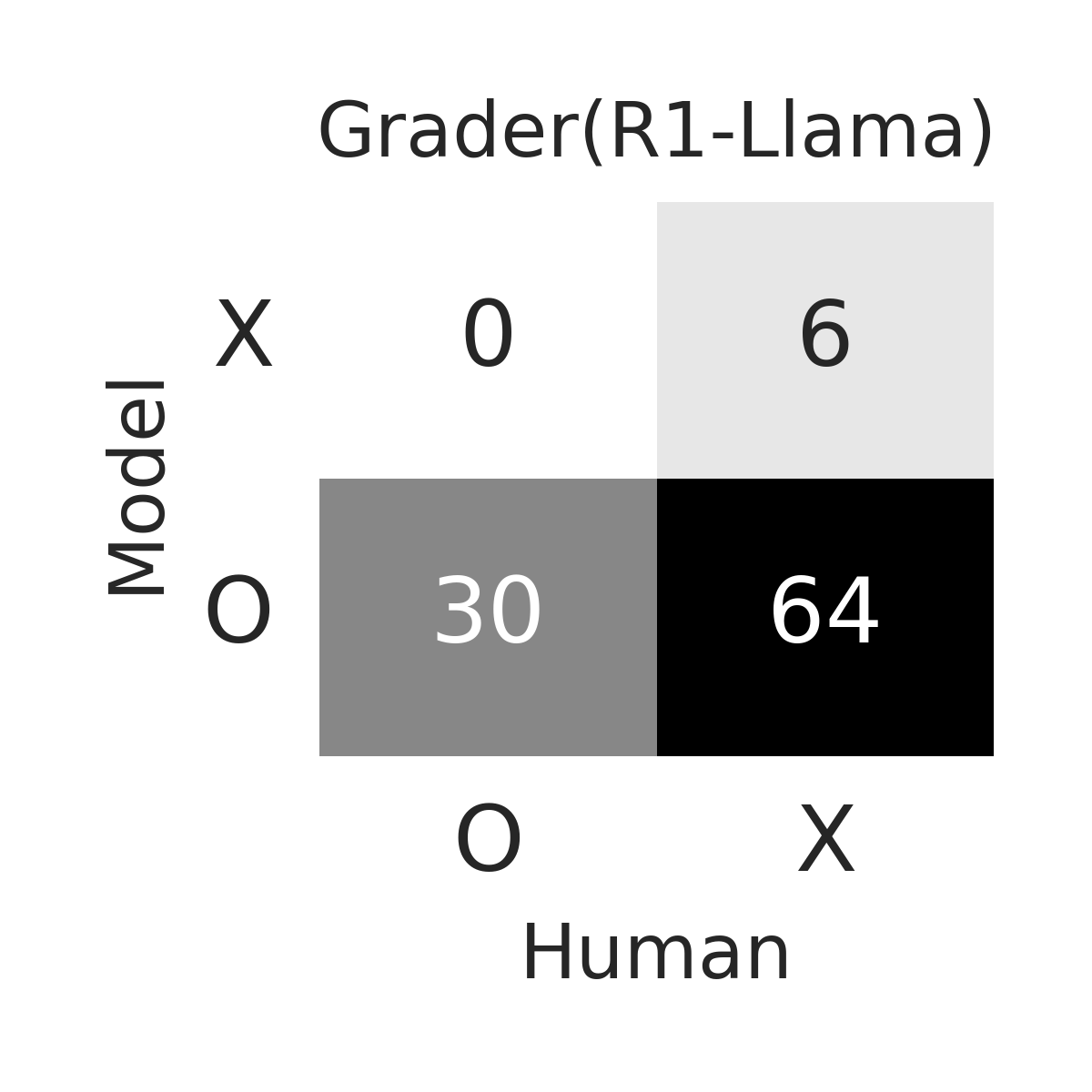}
    \includegraphics[scale=0.42, valign=t,trim={0.5cm 0.5cm 0.25cm 0.25cm},clip]{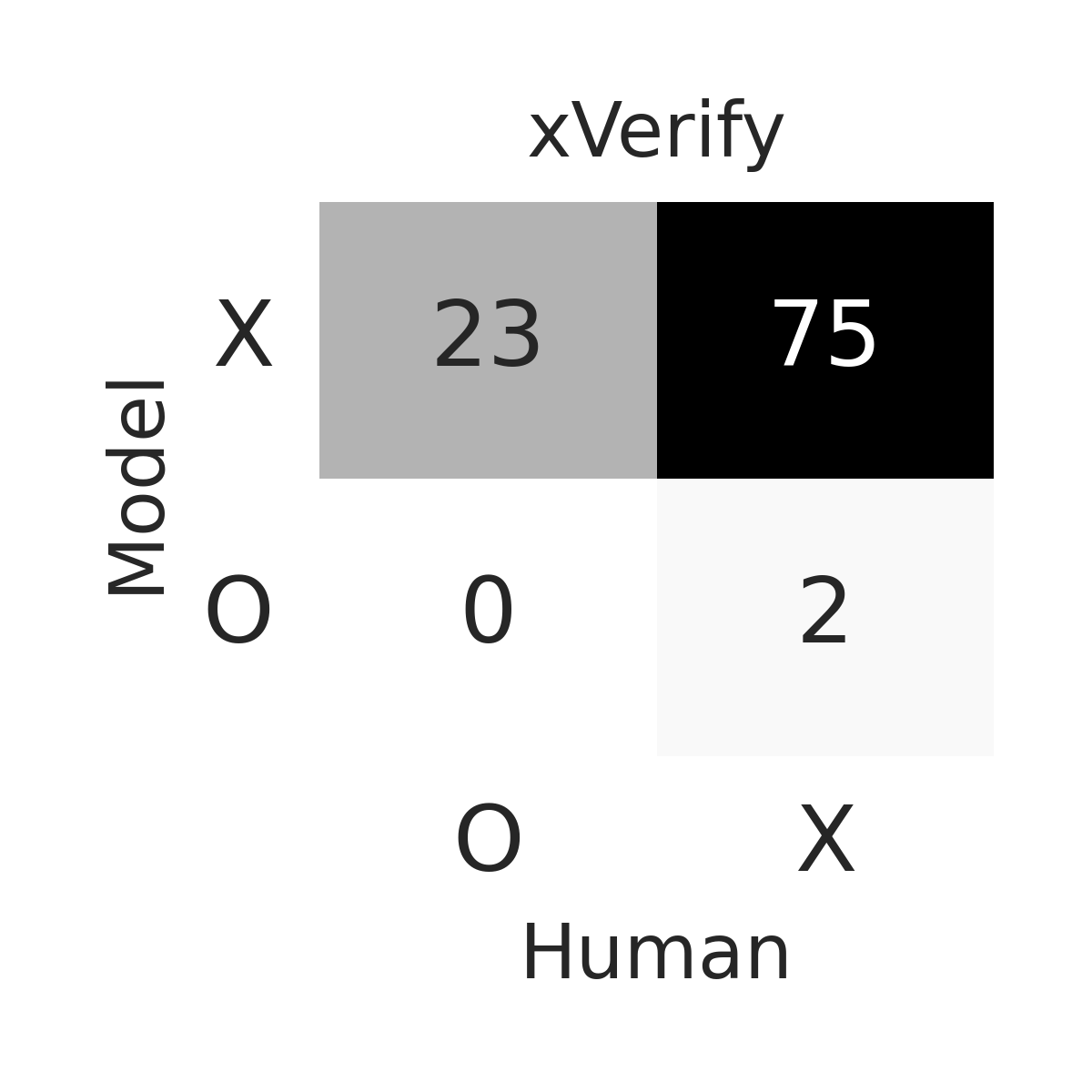}
    \end{subfigure}
    \vspace{-2mm}
    \caption{(left) Performance of reasoning models on open-ended question answering TriviaQA. (right) Confusion matrix illustrating human evaluation performance on 100 samples in determining semantic equivalence between the generated answer and the gold answer. $^*$ denotes the selected results for the detailed human evaluation.}\label{tab:triviaqa-result}
    \vspace{-4mm}
    \end{table*}
    
\subsection{Short-Answer Math Problems}

    We explore the effectiveness of our framework in math domain using GSM8K benchmark~\citep{cobbe2021training}, which features structured (as numbers) but relatively open-ended question answering task.

    As shown in Table~\ref{tab:gsm8k_result}, {\small\tt instructed-format}
    , a template specifically recommended for mathematical problems, performs the best among the various extraction methods. We also modify {\small\tt answer-is-correct} to better handle common mathematical formatting, such as numbers and symbols like \$, ",", and ".". Despite these optimizations, \textsc{Answer Regeneration} with minor post-processing to remove LaTeX commands, such as  \textbackslash boxed\{\} or \textbackslash text\{\}, achieves the highest performance.

    To further validate this, we conduct a human evaluation of instances where the methods' results disagreed. \textsc{Answer Regeneration} framework reports 16.3\% correct, while the conventional answer extraction method is correct in only 6.1\% of the cases. This underscores the superior reliability of our framework even in complex, structured but open-ended domains like mathematics.

\subsection{Open-ended Question Answering}

    Evaluating generative models on open-ended question-answering tasks presents two main challenges: (1) finding the answer within the model's output. (2) determining semantic equivalence between the generated answer and the gold answer. To alleviate the second challenge, we use TriviaQA~\citep{joshi2017triviaqa}, known for its extensive gold answer variations and aliases, minimizing the need for complex semantic matching.

    As shown in Table~\ref{tab:triviaqa-result} (left), \textsc{Answer Regeneration} consistently outperforms direct answer extraction from the reasoning output. We also compare our framework with two other LLM-as-a-judge approaches~\cite{zheng2023judging}: GPTGrader~\citep{wei2024measuring}, which uses an additional inference call with long prompts to categorize semantic similarity into "correct" (same), "incorrect" (not same), and "invalid". xVerify~\citep{chen2025xverify}, a fine-tuned model that evaluates semantic equivalence as "correct" or "incorrect". While the scores from these model-based evaluations might appear better, they carry a critical drawback of model bias.
    Table~\ref{tab:triviaqa-result} (right) presents a human evaluation of semantic equivalence, comparing the model judgement with human judgments on 100 sampled outputs. Qwen3-32B consistently predicts "incorrect" even when the answer is correct and xVerify similarly defaults to "incorrect", and R1-Llama-8B exhibits a bias toward "correct". In contrast, our string-match-based method avoids this model bias and provides a more accurate performance measure, despite of its limitation in determining semantic equivalence.

\section{Discussion and Conclusion}

    Our analysis highlights a critical, yet often overlooked, challenge in evaluating reasoning models: the profound impact of the answer extraction methods on performance scores. We have demonstrated that model performances can fluctuate significantly based on how the final answer is parsed from its reasoning output. This finding suggests that discrepancies between publicly reported scores and reproduced results may stem from undocumented differences not just in prompts, but in the extraction methods itself. To address this issue, we introduced \textsc{Answer Regeneration} framework. Our simple approach offers significant advantages over conventional extraction rules:

    Without specific tuning, the framework consistently achieved superior scores across a variety of tasks, from multiple-choice question answering to short-answer math problems and open-ended question answering tasks. By prompting the model to explicitly state its final answer again, we mitigate inconsistencies caused by diverse output formats.

    Beyond exhibiting better scores than handcrafted, optimized rules, the performance ranking derived from our framework for the Qwen3 model family aligned with the conventional intuition that larger models generally outperform smaller ones. This suggests that the framework provides a more accurate reflection of a model's true capabilities, free from the biases of model-specialized answer extraction rules.
    
    Our method also proves more resilient to common failure of rule-based approach. It successfully handles outputs involving incomplete thinking, models that re-consider their answers, and questions asking for the "incorrect" choice, all of which can confuse rule-based extraction.
    
    Lastly, while LLM-as-a-judge method can suffer from inherent model biases (e.g., consistently predicting, "correct" or "incorrect"), our string-match method, enabled by the concise regenerated output, provides a more reliable  measure of performance.

    In conclusion, through our findings from analysis and the introduction of \textsc{Answer Regeneration} framework, we believe this work contributes toward more reliable and faithful model evaluation for all reasoning-powered LLMs.

    \clearpage
    
\section{Limitations}\label{sec:Limitations}

    \paragraph{Technical Novelty in \textsc{Answer Regeneration}}
    We acknowledge that \textsc{Answer Regeneration} framework itself lacks technical novelty. However, we contend that the value of our contribution lies in the simplicity and the clarity of the results and analysis it provides. Our work demonstrates the benefits of using this framework as a robust and reliable reference for evaluating and fairly comparing the performance of reasoning models.
    
    \paragraph{Experiments with Sophisticated Extraction Rules}
    Our experiments adopted established answer extraction rules from lm-evaluation-harness (\texttt{\small strict-match}, \texttt{\small flexible-match}). Building upon these, we developed more complex, heuristic rules (\texttt{\small answer-is-correct}, \texttt{\small last-extract}) and included the recommended rule for Qwen3 families (\texttt{\small instructed-format}). While we recognize that more aggressively optimized, domain-specific rules could exist, we maintain that such highly-specified rules will still fail to handle the full spectrum of answer variations.

    \paragraph{Experiments with Diverse LLMs and Prompts}
    Our focus was on output-level results, which means that the effect of different input prompts seem to be overlooked. Furthermore, our investigation was limited to publicly available open-source reasoning models. Although greater diversity in models and prompts would enhance generalizability, we believe that the widely-used models and default prompts from established repositories provide sufficiently general results for our findings. We defer the investigation of commercial LLMs, such as ChatGPT, Gemini, and Claude, to future work. As a minor note, we observed that small variations in the input prompts (e.g., changes of option labels or the "Answer:" prefix) do not significantly affect performance.

    \paragraph{Inherent Weakness of \textsc{Answer Regeneration}}
    As discussed in Section~\ref{sec:method}, \textsc{Answer Regeneration} carries inherent limitations. Nonetheless, we believe that employing the simplest possible framework was the most effective way to demonstrate the core benefits of our approach. Exploring further techniques within this framework, such as incorporating concepts like self-consistency~\citep{wang2022self}, represents a valuable direction for future research.

    
\bibliography{anthology,custom}
\bibliographystyle{acl_natbib}

\clearpage
\appendix

\section{Appendix}
    \subsection{Evaluation Toolkits}\label{appendix:related_work}
    
    {\bf MMLU Hendrycks}~\citep{hendryckstest2021} and follow-ups such as MMLU-Pro~\citep{wang2024mmlu} are deeply integrated into most of toolkits, but the original implementation only supports probability based answering for multiple choice question answering. {\bf HELM} (Holistic Evaluation of Language Models; \citet{liang2023holisticevaluationlanguagemodels}) simply use Quasi-exact match that post-process the model generation, such as lower-casing, removing whitespace and punctuation and articles. Also, {\bf OpenCompass}~\citep{2023opencompass} supports both option-likelihood scoring and post-processing option (but provided with blank) to be customized for reasoning outputs, its metrics mainly rely on model-based scoring. Similarly, {\bf lighteval}~\citep{lighteval} has metrics for generated outputs, but there is only a scoring function, not mentioning about post-processing.
    
    \subsection{Regular Expressions used in the Experiments}\label{appendix:exact_regular_expression}
    Note that () makes groups in regular expression and \textbackslash \textbackslash is required both for meta characters and escape sequence in lm-evaluation-harness.
    \begin{itemize}[noitemsep,topsep=1pt]
        \item {\tt strict-match}: ((?<=The answer is )(.*)(?=.)|(?<=answer is )(.*)(?=.)|(?<=The answer: )(.*)(?=.)|(?<=The final answer: )(.*)(?=.))        
        \item {\tt flexible-extract}: (\textbackslash \textbackslash([A-D]\textbackslash \textbackslash))
        \item {\tt instructed-format}:[Aa]nswer\textbackslash"?:\textbackslash \textbackslash s* \textbackslash "?\textbackslash \textbackslash (?([A-D])\textbackslash "|\textbackslash "?\textbackslash \textbackslash **(?([A-D])\textbackslash "
        \item {\tt answer-is-correct}: \textbackslash \textbackslash**[Aa]nswer:\textbackslash \textbackslash**\textbackslash \textbackslash s*(\textbackslash \textbackslash(?[A-D]\textbackslash \textbackslash)?)|\textbackslash \textbackslash**[Aa]nswer\textbackslash \textbackslash**:\textbackslash \textbackslash s*(\textbackslash \textbackslash(?[A-D]\textbackslash \textbackslash)?)|[Aa]nswer is \textbackslash \textbackslash**(\textbackslash \textbackslash(?[A-D]\textbackslash \textbackslash)?)\textbackslash \textbackslash**|[Aa]nswer should be \textbackslash \textbackslash**(\textbackslash \textbackslash(?[A-D]\textbackslash \textbackslash)?)\textbackslash \textbackslash**|[Aa]nswer:\textbackslash \textbackslash s+\textbackslash \textbackslash**(\textbackslash \textbackslash(?[A-D]\textbackslash \textbackslash)?)\textbackslash \textbackslash**|correct answer is \textbackslash \textbackslash**(\textbackslash \textbackslash(?[A-D]\textbackslash \textbackslash)?)\textbackslash \textbackslash**|correct answer:\textbackslash \textbackslash s+\textbackslash \textbackslash**(\textbackslash \textbackslash(?[A-D]\textbackslash \textbackslash)?)\textbackslash \textbackslash**|\textbackslash \textbackslash**(\textbackslash \textbackslash(?[A-D]\textbackslash \textbackslash)?)\textbackslash \textbackslash** is correct|\\**(\textbackslash \textbackslash(?[A-D]\textbackslash \textbackslash)?)\textbackslash \textbackslash** is the correct|\textbackslash \textbackslash**(\textbackslash \textbackslash(?[A-D]\textbackslash \textbackslash)?)\textbackslash \textbackslash** is the answer|\textbackslash \textbackslash**(\textbackslash \textbackslash(?[A-D]\textbackslash \textbackslash)?)\textbackslash \textbackslash** should be the answer
        \item {\tt last-extract}: [\textasciicircum a-zA-Z0-9]([A-D])[\textasciicircum a-zA-Z0-9]
    \end{itemize}
    
    \subsection{Preliminary: Non-reasoning vs. Reasoning}

    \begin{table*}[t]\centering\small
    \scalebox{1.0}{
    \begin{tabular}{lccccc}
    \toprule
    & {\tt Qwen3-32B} & {\tt Qwen3-14B} & {\tt Qwen3-8B} & {\tt R1-Llama-8B} & {\tt R1-Qwen3-8B} \\
    \midrule
    {\tt non-Reason} & 78.4 & 75.7 & 72.2 & 53.0 & 66.2 \\
    {\tt Reason} & 82.1 & 83.8 & 82.1 & 64.8 & 77.6 \\
    \midrule
    {\tt Diff} & +3.7 & +8.1 & +9.9 & +11.8 & +11.4 \\
    \bottomrule
    \end{tabular}
    }
    \caption{The performance comparison when using non-reasoning mode and reasoning mode in LLMs. Non-reasoning mode follows conventional loglikelihood measurements using candidate whereas reasoning mode uses answer extraction algorithms to find the final answer in the reasoning output. The best performance with answer extraction methods are reported.}\label{tab:preliminary_reason}
    \end{table*}

    We demonstrate the power of reasoning in solving MMLU, as presented in Table~\ref{tab:preliminary_reason}. The performance shows that the reasoning significantly improves the model performances. This encourage us to use reasoning model not only for complex problem. but for knowledge-based problems.

    \subsection{Regnerator Independency}

    Table~\ref{tab:llm-as-judge} reports the performance when using different models from the answer generator. Although we use smaller models for regenerator, the performance is similar when using the identical model.

    \begin{table*}[t]\centering\small
    \scalebox{1.0}{
        \begin{tabular}{@{}
        l@{\hskip 0.18cm}
        p{1.4cm}@{\hskip 0.18cm}p{1.4cm}@{\hskip 0.18cm}p{1.4cm}@{\hskip 0.18cm}p{1.7cm}@{\hskip 0.18cm}p{1.7cm}}
        \toprule
        ($\downarrow$) Regenerator & {\tt Qwen3-32B} & {\tt Qwen3-14B} & {\tt Qwen3-8B} & {\tt R1-Llama-8B} & {\tt R1-Qwen3-8B} \\
        \midrule
        {\tt gemma-3-1b-it} & 86.9 & 84.9 & 82.5 & 67.5 & 80.3 \\
        {\tt llama-3.2-1b-it} & 86.5 & 84.4 & 81.8 & 67.8 & 79.6 \\
        {\tt Qwen3-0.6b} & 86.3 & 84.4 & 82.3 & 68.9 & 79.9 \\
        \midrule
        {\tt Qwen3-32B} & \bf 87.1 & 85.2 & 83.7 & 72.1 & 82.6 \\
        {\tt Qwen3-14B} & 87.1 & \bf 85.0 & 83.1 & 71.2 & 81.6 \\
        {\tt Qwen3-8B} & 87.4 & 85.2 & \bf 83.3 & 72.5 & 82.0 \\
        {\tt R1-Llama-8B} & 87.0 & 84.9 & 82.6 & \bf 68.8 & 80.2 \\
        {\tt R1-Qwen3-8B} & 84.2 & 81.0 & 81.1 & 70.9 & \bf 80.7 \\
        \bottomrule
        \end{tabular}
    }
    \caption{Model performance when different models are used for \textsc{Answer Regeneration} step. Bold indicates the reported score when the reasoning models and the regenerators are the same.}\label{tab:llm-as-judge}
    \end{table*}

\end{document}